\documentclass[10pt,twocolumn,letterpaper]{article}

\usepackage{cvpr}              

\definecolor{mygreen}{RGB}{34,139,34}  
\definecolor{mypurple}{RGB}{150, 50, 255}
\definecolor{myred}{RGB}{180, 0, 0}
\definecolor{mygreen}{RGB}{0, 150, 0}
\definecolor{mybrown}{RGB}{150, 75, 0}
\definecolor{mypink}{RGB}{200, 50, 150}
\definecolor{mygold}{RGB}{200, 150, 0}
\definecolor{lightpurple}{RGB}{220, 200, 255}
\definecolor{myblue_fill}{RGB}{195,218,255}
\definecolor{myblue_shape}{RGB}{31,128,150}
\definecolor{myblue}{RGB}{21,34,137}

\usepackage{xcolor}
\usepackage{tcolorbox}
\usepackage{amsmath}
\usepackage{multirow}
\usepackage{longtable}
\usepackage{ltablex}  
\usepackage{booktabs} 
\usepackage{subcaption} 
\usepackage{ulem}
\usepackage{float}
\usepackage{fancyvrb}
\usepackage[accsupp]{axessibility}

\definecolor{cvprblue}{rgb}{0.21,0.49,0.74}
\usepackage[pagebackref,breaklinks,colorlinks,allcolors=cvprblue]{hyperref}

\def\paperID{26} 
\def\confName{CVPR}
\def\confYear{2025}

\title{Visually Interpretable Subtask Reasoning for Visual Question Answering}

\author{
Yu Cheng\textsuperscript{1} \quad
Arushi Goel \textsuperscript{2} \quad
Hakan Bilen \textsuperscript{1} \\
\textsuperscript{1}University of Edinburgh\quad
\textsuperscript{2}NVIDIA \\
{\tt\small s2521923@ed.ac.uk, goel.arushi@gmail.com, hbilen@ed.ac.uk}
}

\begin{document}

\twocolumn[{%
    \renewcommand\twocolumn[1][]{#1}%
    \maketitle
    \centering
    \includegraphics[width=0.9\linewidth]{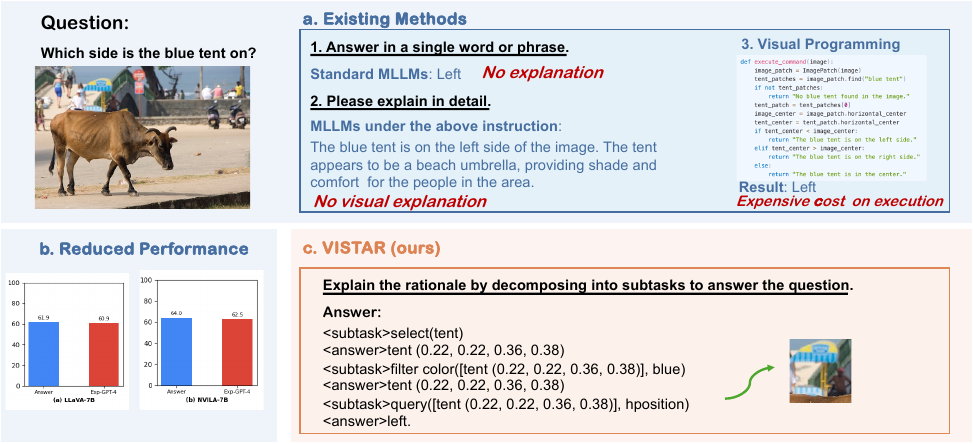}
    \captionof{figure}{\textbf{Comparison of standard MLLMs, programmatic reasoning and VISTAR across the output.}(a) Existing methods either provide no explanations, lack visual grounding, or have high computational costs. (b) Performance degradation of standard MLLMs (LLaVA-1.5-7B~\citep{Liu2023ImprovedBW} and NVILA-8B~\citep{Liu2024NVILAEF}) when forced to generate explanations alongside answers. `Exp-GPT-4' evaluates semantic similarity using \textit{GPT-4-turbo}~\citep{Achiam2023GPT4TR}. (c) VISTAR  effectively addresses these issues by decomposing the question into structured sub-tasks, providing both visual (bounding box, following the format ($x_l, y_l, x_r, y_r$) where $x_l, y_l$ are the coordinates of the top-left corner and $x_r, y_r$ are the coordinates of the bottom-right corner) and textual rationales without compromising accuracy.}
    \label{fig:splash}
    }
    ]

\begin{abstract}

Answering complex visual questions like `Which red furniture can be used for sitting?' requires multi-step reasoning, including object recognition, attribute filtering, and relational understanding. Recent work improves interpretability in multimodal large language models (MLLMs) by decomposing tasks into sub-task programs, but these methods are computationally expensive and less accurate due to poor adaptation to target data.
To address this, we introduce VISTAR (Visually Interpretable Subtask-Aware Reasoning Model), a subtask-driven training framework that enhances both interpretability and reasoning by generating textual and visual explanations within MLLMs. Instead of relying on external models, VISTAR fine-tunes MLLMs to produce structured Subtask-of-Thought rationales (step-by-step reasoning sequences). Experiments on two benchmarks show that VISTAR consistently improves reasoning accuracy while maintaining interpretability. Our code and dataset will be available at \url{https://github.com/ChengJade/VISTAR}.

\end{abstract}    

\section{Introduction}
\label{sec:intro}

Recent advancements in multimodal large language models (MLLMs) have led to significant performance improvements on image-text paired tasks like image captioning~\citep{Zhang2024Ferretv2AI, Goncharova2024OmniFusionTR}, visual question answering (VQA)~\cite{liu2023visual, Liu2023ImprovedBW}, and video understanding~\cite{Huang2023VTimeLLMEL, Kalarani2024SeeingTU}. Despite these advancements, a critical limitation of current MLLMs is their lack of interpretability. 
Notably, in terms of VQA task, recent MLLMs~\cite{Liu2023ImprovedBW, Liu2024NVILAEF, Wang2024Emu3NP} are primarily trained under instructions to generate direct answers, optimizing for correctness rather than interpretability. 
When these models are prompted to generate explanations alongside their predictions, their answer accuracy declines significantly. 
This phenomenon arises because current MLLMs are not explicitly trained to engage in stepwise reasoning, and introducing interpretability often interferes with their optimization for final answer prediction. 
Furthermore, most existing models primarily generate textual explanations, lacking visual justifications such as object-level grounding, which is essential for human-aligned reasoning in the VQA task.

To enhance interpretability in the VQA task, recent approaches such as VISPROG~\citep{Gupta2022VisualPC} and ViperGPT~\citep{Suris2023ViperGPTVI} have proposed decomposing complex queries into a sequence of sub-tasks in modular programming languages, where each step is executed by a pre-trained model (\eg, object detector, pose detector), thereby making the reasoning process more transparent.
However these approaches have multiple shortcomings.
First, executing multiple models sequentially incurs high computational costs. Second, they rely on pre-trained models that may have been trained on disparate data distributions, operating in a zero-shot setting without parameter adaptation, which can lead to performance drops. Third, these methods are inherently error-prone, as the generated programs may omit necessary steps, introduce spurious operations, or fail to recover from errors in intermediate outputs.
To mitigate the last limitation, VPD~\citep{Hu2023VisualPD} introduces an instruction-tuning framework that distills the reasoning capabilities of large language models (LLMs) into MLLMs, allowing them to sample multiple candidate programs and identify the correct one. 
Despite improving answer accuracy, VPD still relies on program execution thereby inheriting the first two limitations.

Building upon the limitations of the prior work, we introduce \textbf{Visually Interpretable Subtask-Aware Reasoning Model (VISTAR)}, a framework designed to enhance the interpretability and reasoning of MLLMs in VQA. 
To highlight the key differences between standard MLLMs and our approach, Fig.~\ref{fig:splash} presents two examples where both models correctly predict the answer. Our method goes beyond direct answer generation by providing structured reasoning steps, including a textual explanation that decomposes into sub-tasks and object-level visual grounding via bounding boxes. Specifically, VISTAR introduces LLM-prompted \textbf{Subtask-of-Thought (SoT)} rationales, which decompose visual queries into structured sub-tasks such as object selection, attribute filtering, and relationship verification while generating intermediate answers in the form of object names with bounding boxes, categorical attributes, or binary decisions. 
Next, VISTAR employs a multi-modal instruction tuning strategy to integrate these structured SoT rationales into the learning process, enabling MLLMs to generate both interpretable reasoning steps and accurate answers. 
Unlike existing interpretable methods that rely on computationally expensive program execution, VISTAR learns structured reasoning directly within an MLLM, eliminating the need for external program execution while preserving interpretability.

Hence, the main contributions of this work can be summarized as follows:
\begin{itemize}
    \item We introduce SoT, an LLM-generated dataset that structures visual queries into step-by-step reasoning sub-tasks.
    \item We propose VISTAR, an instruction tuning pipeline that enables MLLMs to generate visual (object-level bounding boxes) and textual (sub-task reasoning process) explanations.
    \item Experimental results demonstrate that VISTAR achieves improved interpretability and reasoning accuracy, and generalizes well to datasets in the zero-shot setting.
\end{itemize}

\section{Related work}

\paragraph{Multimodal Large Language Models (MLLMs).}
With the advancement of LLMs in text understanding and reasoning—-such as GPT~\citep{radford2019language, Brown2020LanguageMA}, Vicuna~\citep{vicuna2023}, LLaMA~\citep{Touvron2023LLaMAOA}, and Qwen~\citep{Bai2023QwenTR}, there is growing demand in developing MLLMs that extend these capabilities to both text and images. Examples include BLIP2~\citep{Li2023BLIP2BL}, InstructBLIP~\citep{instructblip}, MiniGPT-4~\citep{Zhu2023MiniGPT4EV}, Qwen-VL~\citep{Bai2023QwenTR, Wang2024Qwen2VLEV}, LLaVA~\citep{liu2023visual, Liu2023ImprovedBW}, and NVILA~\citep{Liu2024NVILAEF}.

MLLMs typically use a pre-trained visual encoder (\eg, CLIP~\citep{Radford2021LearningTV} or SigLIP~\citep{Zhai2023SigmoidLF}) to extract visual features, which are then mapped into textual token embeddings via a connector before being processed by an LLM. Training involves two key stages: large-scale image-text pair alignment to synchronize visual and textual representations, followed by instruction tuning to enhance task adaptability. 
The visual encoder is usually frozen, while the connector and LLM weights are updated during instruction tuning.
These models are commonly evaluated on various vision-language tasks~\citep{Huang2023VTimeLLMEL, ma2024groma, Dong2024InsightVEL, zhang2025mllms}, including VQA~\citep{zhang2025mllms}, image captioning~\citep{Zhang2024Ferretv2AI}, visual grounding~\citep{ma2024groma}, and multimodal reasoning~\citep{Dong2024InsightVEL}. Among them, VQA~\citep{Li2024ASO} is a crucial benchmark for assessing MLLM's ability to understand and reason over visual content by answering questions about images. However, most existing MLLMs generate only brief textual responses, lacking interpretability~\citep{Liu2023ImprovedBW, Liu2024NVILAEF}.

\paragraph{Interpretability.}
Most existing MLLMs, such as LLaVA~\citep{liu2023visual, Liu2023ImprovedBW}, are primarily trained on VQA tasks with instructions like ``answer the question in a single word or phrase''. While this ensures concise responses, it restricts the model's ability to provide explanations. Although prompt tuning (\eg, ``give a more detailed explanation") can elicit longer explanations, the generated explanations often lack depth and explicit visual reasoning, limiting overall interpretability~\citep{lou2025sae}.

Recent research has focused on improving MLLM interpretability in VQA through modular programs that break down visual queries into Python-like components~\citep{Suris2023ViperGPTVI, Gupta2022VisualPC, ke2024hydra}. 
These methods leverage the in-context learning ability of code-generation model~\citep{chen2021evaluating} to dynamically assemble vision-language modules, enabling explicit reasoning and better generalization. By executing generated programs using predefined APIs, these frameworks eliminate the need for additional training. 
However, this approach relies on multiple pre-trained models—such as object detection~\citep{Li2021GroundedLP}, visual question answering~\citep{Li2023BLIP2BL}, depth estimation~\citep{Ranftl2019TowardsRM} etc. —which significantly increases computational costs. 

Beyond modular programs, another promising direction for enhancing the reasoning and interpretability of MLLMs in VQA is question decomposition~\citep{khan2023exploring, Ma2024TaskND,Zhang2024VisualQD}. This method breaks down complex queries into sub-questions, effectively introducing a new form of chain-of-thought (CoT) reasoning that allows models to reason step by step. Recent works~\citep{Ma2024TaskND, Zhang2024VisualQD} have shown that question decomposition not only improves interpretability but also enhances accuracy by aligning with human reasoning strategies. Some approaches enable MLLMs to generate their own decompositions through in-context learning~\citep{khan2023exploring}, while others propose fine-tuning models using LLM-generated question decomposition data~\citep{Ma2024TaskND, Zhang2024VisualQD}. 
However, question decomposition is not always beneficial for interpretability, since it is also unable to provide visual explanations.

Most related to ours, Hu et al.~\citep{Hu2023VisualPD} converted generated programs into CoT reasoning and distilled the resulting data into MLLMs. While this approach enhances reasoning capabilities, it remains reliant on program execution of pre-trained models, making it computationally intensive and failing to adapt them to the target data.
In contrast, our model is directly trained for the target task while being computationally efficient and providing accurate answers along with textual and visual explanations.

\section{Method}

\begin{figure*}[!ht]
    \centering
    \includegraphics[width=0.7\textwidth]{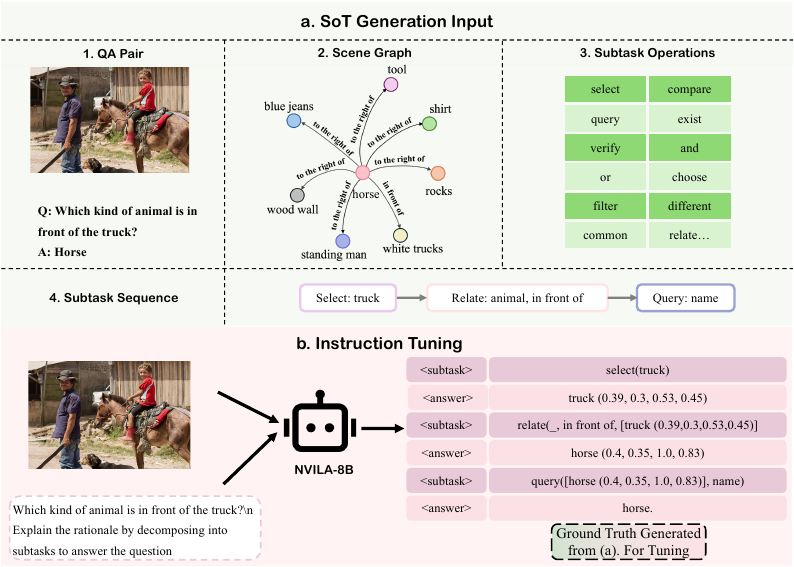} 
    \caption{\textbf{Overview of VISATR.} VISTAR uses an LLM to generate faithful SoTs via in-context learning. Given the input including the query, ground truth, scene graph in an image and sub-task operation sequence inside the dataset (top), LLaMA-3.1-70B-Instruct would output a SoT to answer the query. The generated SoT is then used to fine-tune an MLLM, enabling it to produce both visual (object-level bounding boxes) and textual explanations during inference (bottom).}
    \label{fig:VISTAR}
\end{figure*}

Our approach, VISTAR leverages LLM-generated SoT rationales, which decomposes the VQA task into structured intermediate reasoning steps, making the model’s decision-making process more transparent and interpretable (see Fig.~\ref{fig:VISTAR}). 
Here, we first present the pipeline for generating SoT rationales, detailing how these intermediate steps contribute to multimodal understanding. 
We then describe our training strategy, which incorporates these rationales into the learning process, enabling MLLMs to better utilize structured reasoning for improved interpretability and performance.

\subsection{Problem Formulation}
Given an image $I$ and a question $Q$, our goal is to generate SoT rationales-- a structured reasoning process that decomposes the question into a sequence of sub-tasks, each producing intermediate results that are required to arrive at the correct answer along with the final answer. 
A SoT $S$ (illustrated in Fig.~\ref{fig:splash}.(c)) is defined as an ordered set of reasoning steps:
\begin{equation}
S = \{ (\texttt{op}_1, r_1), (\texttt{op}_2, r_2), ..., (\texttt{op}_n, r_n) \}
\label{eq:Sopr}    
\end{equation} where $\texttt{op}_i$ represents the $i$-th sub-task operation, such as selecting an object, filtering color, or querying an attribute, and $r_i$ is its corresponding intermediate result and $n$ is the total number of operations. 
The final result, $r_n$, is the predicted answer to the question.
Each intermediate result $r$ consists of $\left\{a, b\right\}$ where $a$ is the intermediate textual answer and $b=(x_l,y_l,x_r,y_r)$ is the bounding box representation, a four-dimensional tuple containing top left and bottom right coordinates, normalized by dividing each coordinate by the larger dimension of the image (either width or height).
Our goal is to learn a function $f_{\theta}: (I, Q) \rightarrow S
$, an MLLM model parameterized by $\theta$ that maps a pair $(I,Q)$ to $S$, the SoT sequence.

\subsection{SoT generation}\label{subsec:data generation}
Here, we explain our pipeline for synthesizing SoT for the VQA task (see in the top part of Fig.~\ref{fig:VISTAR}) on the GQA dataset~\citep{Hudson2019GQAAN}. 

Given an image $I$ with $(Q, SG, \{\texttt{op}_i\}_{i=1}^N, y)$—where $Q$ is a textual query, $SG$ is the scene graph of the image encoding objects, attributes, and relationships between objects, $\texttt{op}$ is the set of sub-task operations, and $y$ is the ground truth answer for the $Q$ query, we obtain the corresponding SoT by predicting the missing intermediate answers $\{r_i\}_{i=1}^{N-1}$ where $r_N=y$ through the following steps:

\noindent\textbf{1. Prompt-based LLM generation}: 
As manually labeling the intermediate answers is expensive, we automatically obtain the SoT.
As the given $SG$ includes all the required visual information, we use an LLM~\citep{Zhu2023MiniGPT4EV,Touvron2023LLaMAOA} (not an MLLM) for this step by
prompting it with a structured prompt consisting of (see Appendix \S\ref{sec:prompt_for_sot} for more details):

\begin{itemize}
    \item $Q$: The original textual query from the dataset.
    \item \textbf{The ground truth answer} $y$: To guide the expected reasoning process.
    \item $SG$: A structured representation of objects, attributes, and relationships.
    \item $\{\texttt{op}_i\}_{i=1}^N$: The GQA dataset defines 145 sub-task operations in a specific format (Fig.\ref{fig:subtask_example}). To improve clarity and interpretability, we reformat these operations within our prompt, ensuring the LLM produces outputs in a structured SoT format (Fig.\ref{fig:subtask_example}). Each sub-task operation functions as a predefined function with a fixed set of arguments (see Appendix \S\ref{sec:subtask_op_def}). 
    \item \textbf{In-context Learning}: In the prompt, we illustrate the SoT generation through a few examples to get desirable outcomes.
\end{itemize}

Given this structured input, LLM can generate a SoT in the format of \cref{eq:Sopr}.
In particular, the outputs of each sub-task operation $r_i$ can be objects with their bounding boxes, categorical attributes (\eg, `red'), or binary decisions (`yes'/`no'). For instance, as shown in Appendix \S\ref{sec:example_of_sot_op} Fig.~\ref{fig:subtask_example}, given a complex query, the model produces the SoT as:

\begin{itemize}
    \item \texttt{<subtask>select(garland)}: The model first identifies the target object (\texttt{garland}) and outputs its corresponding intermediate answer: \texttt{<answer>garland <bbox>(0.51, 0.0, 0.54, 0.09)}.
    \item \texttt{<subtask>relate(curtain, to the right of, [garland <bbox>(0.51, 0.0, 0.54, 0.09)])}: Next, the model determines which object satisfies the specified spatial relationship, identifying \texttt{curtain} as being ``to the right of" the garland. The resulting answer is \texttt{<answer>curtain <bbox>(0.73, 0.0, 0.87, 0.58)}.
    \item \texttt{<subtask>relate(furniture, same color, [curtain <bbox>(0.73, 0.0, 0.87, 0.58)])}: The model then finds a piece of furniture that shares the same color as the \texttt{curtain}, producing the result \texttt{<answer>couch <bbox>(0.12, 0.48, 0.71, 0.97)}.
    \item \texttt{<subtask>query([couch <bbox>(0.12, 0.48, 0.71, 0.97)], name)}: Finally, the model retrieves the name of the identified furniture, yielding the answer \texttt{<answer>couch}.
\end{itemize}

\noindent \textbf{2. SoT filtration}: Since LLM-generated data is prone to errors, we remove the noisy SoTs when their final answer does not align with the ground truth answers, they do not conform to the expected argument format, and their length exceeds a specified threshold  to prevent excessively long reasoning.

\subsection{Instruction tuning for interpretability}
Here, we present the details for training using the data generated in \S\ref{subsec:data generation}, with the goal of providing both visual and textual explanations to support MLLM answer prediction. Building on the fine-tuning of MLLMs for instruction-based tasks, we employ a visual instruction tuning approach that aligns the model’s reasoning process with SoT rationales. Specifically, we prepend a task-specific instruction ``Explain the rationale by decomposing into subtasks to answer the question'', to each query $Q$. Given a training sample $(I, Q, S)$, we fine-tune the MLLM $f$ to minimize the loss associated with predicting the correct SoT, $S$, including the final answer. The resulting optimization is as follows:
\begin{equation}
\begin{aligned}
\min_{\theta} \sum_{j=1}^N \ell\left(f_{\theta}\left(I_j, Q_j\right), S_j\right)
\end{aligned}
\end{equation}
where $I$ represents the input image, $Q$ is the textual query appended with our instruction, and $S$ denotes the structured SoT reasoning. The function $\ell(\cdot)$ represents the next-token prediction loss, and $N$ is the total number of training samples. This objective encourages the model to generate well-structured reasoning sequences that align with human-interpretable step-by-step explanations. During training, we initialize from a pre-trained MLLM and fine-tune it using our generated SoT dataset. The vision encoder remains frozen, while we update the visual adaptor and the language model to effectively integrate both textual and visual explanations. By leveraging instruction tuning, we enhance the model’s interpretability and generalization, enabling it to provide more structured and explainable responses for the VQA task, even under the zero-shot setting.

\section{GQA-SoT}
\label{sec:gqasot}
We apply our SoT generation method on the GQA dataset~\citep{Hudson2019GQAAN} that originally consists of $113K$ images with $22M$ question-answer pairs along with sub-task operation sequences (see Table~\ref{tab:dataset_statistics}).
For SoT generation, we retain all training images but selectively sample questions to ensure balanced coverage of all question types and allow our model to learn diverse reasoning patterns while avoiding redundant data generation and excessive computational costs associated with annotating the entire training set. 
As a result, we generate $249K$ SoT annotations on the GQA dataset~\citep{Hudson2019GQAAN} for fine-tuning, and $132K$ for validation ensuring quality evaluation of predicted SoT from the fine-tuned model (as shown in Table~\ref{tab:dataset_statistics}). 
Notably, as there is no ground truth of scene graph for the test set, we are unable to generate rationales for the test set.
We call this SoT augmented dataset, \emph{GQA-SoT}.

\begin{table}[h]
    \centering
    \caption{\textbf{Statistics of our LLM-generated SoT annotations used for training and evaluation on GQA.}}
    \label{tab:dataset_statistics}
    \scalebox{0.8}{
    \begin{tabular}{l|c|c|c|c}
        \toprule
        \textbf{Dataset} & \textbf{Split} &  \textbf{Images} &  \textbf{\# QA Pairs} & \textbf{\# SoTs} \\
        \midrule
        \multirow{4}{*}{GQA~\citep{Hudson2019GQAAN}} 
        & Train      & 90.5K & 14.3M & - \\
        & Validation & 8.1K  & 132K  & -\\
        & Test       & 3.0K  & 1.51M & - \\
        & others     & 16.59K & 4.95M  & -\\
        \midrule
        \multirow{3}{*}{GQA-SoT(Ours)} 
        & Train-balanced     & 77.4K & 994K  & 249K \\
        & Validation         & 8.1K  & 132K  & 132K\\
        & Test-dev-balanced  & 0.3K  & 12.5K & - \\
        \bottomrule
    \end{tabular}}
\end{table}

\section{Experiments}
Here, we evaluate the effectiveness of our approach in enhancing the interpretability of MLLMs for VQA. 
Specifically, we finetune \textit{NVILA-8B} using our instruction tuning data, a state-of-the-art MLLM known for its strong performance in vision-language tasks. 

This section describes our experimental setup (\S\ref{subsec:setup}), assesses compositional reasoning in VQA task rigorously (\S\ref{subsec:compo_vqa}),  interpretability of our model (\S\ref{subsec:interpretability}) and ablate various design decisions (\S\ref{subsec:ablation}).

\subsection{Experimental Setup}\label{subsec:setup}
\noindent\textbf{Backbone model. }We perform experiments with \textit{NVILA-8B}~\citep{Liu2024NVILAEF} as our base model which is a strong baseline with competitive performance on both image and video benchmarks, and is equipped with strong multimodal reasoning capabilities.

\noindent\textbf{Datasets. } The generation of SoT relies on extracting sub-task operations from the GQA dataset~\citep{Hudson2019GQAAN} as discussed in \S\ref{sec:gqasot}, where other VQA datasets are unavailable for these sub-task operations. 
Hence, we fine-tune our model exclusively on the GQA-SoT and evaluate its performance on the same dataset.

We also evaluate our model on the CRIC dataset~\citep{gao2023cric}, after sub-selecting questions that require complex reasoning steps from its evaluation set.
Note that this evaluation is performed as zero-shot without training our model on CRIC to assess our model’s generalization ability beyond the GQA dataset while focusing on compositional reasoning in unseen scenarios.

\noindent\textbf{Training setup. }We fully fine-tune the visual adaptor and the LLM backbone of the \textit{NVILA-8B}~\citep{Liu2024NVILAEF} initialized model using 8-A100 NVIDIA GPUs. Specifically, we use a learning rate of 1.5e-5 with a warmup ratio of 0.3 and a cosine decay, a global batch size of 128, and a total of 1 training epochs. The training process takes approximately 4 hours.

\noindent\textbf{Baselines. }We compare our proposed method to baselines for the final answer accuracy on the test set of the GQA dataset~\citep{Hudson2019GQAAN}. 
Importantly, we compare to the original \textit{NVILA-8B} model~\citep{Liu2024NVILAEF}, other recent MLLMs including \textit{MiniGPT-4}~\citep{Zhu2023MiniGPT4EV}, \textit{InstructBLIP}~\citep{instructblip}, \textit{Qwen-VL-Chat}~\citep{Bai2023QwenTR}, \textit{LLaVA-1.5}~\citep{Liu2023ImprovedBW}, and visual compositional reasoning methods such as \textit{ViperGPT}~\citep{Suris2023ViperGPTVI} and \textit{VPD}~\citep{Hu2023VisualPD}. 
We also assess our model's interpretability in terms of object localization and textual interpretability, and compare our model with \textit{NVILA-8B} (backbone)~\citep{Liu2024NVILAEF}.

\noindent\textbf{Evaluation setup. }
Each SoT consists of the final prediction, sub-task operations and object-level bounding boxes. 
First we evaluate the answer accuracy in the VQA task by comparing the model's final predictions to the ground-truth answers. 
Next, we assess the accuracy of the sub-task operations (\eg, select), focusing on the correctness of the predicted reasoning steps without considering their arguments (\eg, truck). 
This is crucial to determine whether the model follows a logically reasoning process when decomposing complex queries into sub-tasks. Since generating our SoT that contains sub-task operations and their arguments (\eg, select(truck)) is similar to open-ended text generation, we use \textit{GPT-4-turbo} to better capture the similarity to the ground truth. In addition, we also include human evaluation to evaluate the consistency and quality of predicted SoT by our fine-tuning model.

Finally, to verify visual explanations based on the object bounding box, we use the validation set for evaluation (\S\ref{sec:gqasot}) as there is no bounding box provided in the test dataset of GQA~\citep{Hu2023VisualPD}. For VQA-based object grounding evaluation, we adapt widely-used Intersection over Union (IoU)~\citep{Rezatofighi2019GeneralizedIO}, precision, and recall to VQA-based object detection. Specifically, we calculate \textbf{IoU} by averaging the \textit{IoU scores} of \textit{all correct predictions}, considering only cases where the model’s answer is correct. We define \textit{True Positives (TP)} as instances where the predicted answer is correct and the detected bounding box IoU exceeds a given threshold. \textit{False Positives (FP)} include cases 1) where the answer is correct but IoU is below the threshold; 2) the answer is incorrect and IoU is below the threshold; 3) or the answer is incorrect but IoU is above the threshold. The total ground truth samples include all evaluated instances. Given this, \textbf{precision} is computed as the ratio of \textit{TP} to \textit{the sum of TP and FP} while \textbf{recall} is defined as the ratio of \textit{TP} to \textit{the number of all the ground truth samples}.

\subsection{Compositional reasoning visual question answering}\label{subsec:compo_vqa}

We assess our model's ability finetuned on the GQA-SoT dataset to generate answers on benchmarks that require multi-step reasoning and compositional understanding. Table~\ref{tab:gqa_results} shows the accuracy of baseline VQA models on the test set of the GQA dataset~\citep{Hudson2019GQAAN}. It is important to note that the baseline models, except for ViperGPT~\cite{Suris2023ViperGPTVI}, generate the direct answer without providing any explanations. As shown in Table~\ref{tab:gqa_results}, our model achieves a new state-of-the-art performance on the GQA dataset~\citep{Hudson2019GQAAN} while providing step-by-step textual and visual reasoning.
\begin{table}[t]
    \centering
    \caption{\textbf{Comparison of accuracy (\%) on the GQA~\citep{Hudson2019GQAAN} test set.} We report results for various state-of-the-art models.}
    \label{tab:gqa_results}
    \begin{tabular}{ll}
        \toprule
        & Accuracy (\%) $\uparrow$ \\
        \midrule
        ViperGPT (zero-shot)~\citep{Suris2023ViperGPTVI} & 48.1 \\
        MiniGPT-4 (Vicuna-13B)~\citep{Zhu2023MiniGPT4EV} & 43.5  \\
        InstructBLIP (Vicuna-13B)~\citep{instructblip} & 49.5 \\
        Qwen-VL-Chat (9.7B)~\citep{Bai2023QwenTR} & 57.5 \\
        PaLI-3-VPD (5B)~\cite{Hu2023VisualPD} & 61.3 \\
        PaLI-3-VPD (55B)~\citep{Hudson2019GQAAN} & 63.3 \\
        LLaVA-1.5 (Vicuna-7B)~\citep{Liu2023ImprovedBW} & 61.9 \\
        LLaVA-1.5 (Vicuna-13B)~\citep{Liu2023ImprovedBW} & 63.2 \\
        \midrule
        NVILA (8B)~\citep{Liu2024NVILAEF}  & 64.0  \\
        \textbf{VISTAR (fine-tune on NVILA-8B)} & \textbf{65.1} \textcolor{mygreen}{$\uparrow$ (+1.1)} \\
        \bottomrule
    \end{tabular}
    
\end{table}

Additionally, we conduct a zero-shot evaluation on the CRIC~\citep{gao2023cric} dataset, a compositional reasoning and common sense dataset unseen during training, to assess the model's ability to generate structured explanations without supervision. We sampled 10,000 questions from the validation set of CRIC~\citep{gao2023cric} dataset. As CRIC has not been widely tested on other baselines, there are no available results for direct comparison, so we compare directly to our baseline \textit{NVILA-8B}~\citep{Liu2024NVILAEF}. 
In Table~\ref{tab:cric_results}, we observe that our model outperforms the NVILA-8B~\citep{Liu2024NVILAEF} on CRIC~\citep{gao2023cric}, achieving an accuracy of 61.1\% compared to 60.8\%. 
The result suggest that our model not only improves performance on the seen GQA dataset but also generalizes to unseen datasets like CRIC~\citep{gao2023cric}.
\begin{table}[h]
    \centering
    \caption{\textbf{Comparison of accuracy (\%) on the randomly sampled CRIC dataset.} We report the result in a zero-shot setting.}
    \label{tab:cric_results}
    \scalebox{0.8}{
    \begin{tabular}{lc}
        \toprule
        & Accuracy (\%) $\uparrow$ \\
        \midrule
        NVILA (8B zero-shot)~\citep{Liu2024NVILAEF}  &  60.8 \\
        \midrule
        \textbf{VISTAR (8B zero-shot)} & \makebox[1.7cm][r]{\textbf{61.1}} \textcolor{mygreen}{$\uparrow$ (0.3)} \\
        \bottomrule
    \end{tabular}}
    
\end{table}

\subsection{Evaluation on interpretability}\label{subsec:interpretability}
Here, we evaluate the quality of predicted SoT rationales across both visual and textual explanations. 
As mentioned in \S\ref{sec:gqasot}, the SoT for the test set is not available, we evaluate the performance of SoT prediction on a sampled subset of the validation set.

\begin{figure*}[htbp]
    \centering
    \includegraphics[width=0.80\textwidth]{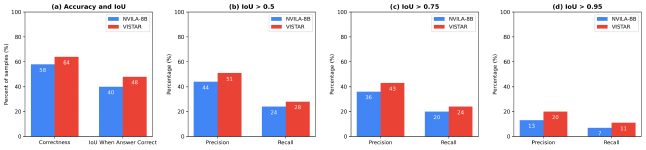} 
    \caption{\textbf{Quantitative results of visual explanations.} (a) Comparison of answer correctness and object localization accuracy (measured by IoU when the answer is correct). (b-d) Precision \& recall at different IoU thresholds (0.5, 0.75, 0.95). VISTAR consistently outperforms NVILA-8B in both accuracy and object-level visual grounding, demonstrating improved interpretability and localization quality.}
    \label{fig:vis_exp_result}
\end{figure*}

\noindent\textbf{Quantitative results of visual explanations. }First, we assess the accuracy of our object-level bounding box predictions. 
Since our backbone model, \textit{NVILA-8B}~\citep{Liu2024NVILAEF}, does not natively generate bounding boxes for the VQA task, we first obtain its predicted answer using the original VQA instruction, \textit{``Answer in a single word or phrase."}. We then append a follow-up instruction from its object grounding task, \textit{``Provide the bounding box coordinate of the region."}, to extract the corresponding bounding box (as shown in Appendix \S\ref{sec:prompt_bbox} Fig.~\ref{fig:instruc_bbox}). For our proposed method \textit{VISTAR}, we extract the operation \texttt{query([object name <bbox> (bounding box)], name)} at the final reasoning step from the SoT. It ensures that the predicted answer corresponds to an object, enabling a direct comparison of bounding boxes between \textit{VISTAR} and \textit{NVILA-8B}~\citep{Liu2024NVILAEF}. 
By aligning the final predicted object names with their respective bounding boxes, we effectively evaluate the object-level localization accuracy of both models. As shown in Fig.~\ref{fig:vis_exp_result}, VISTAR consistently outperforms \textit{NVILA-8B}~\citep{Liu2024NVILAEF} in both answer correctness and bounding box accuracy. Specifically, VISTAR achieves higher accuracy in predicting correct answers while also improving object localization, as evidenced by increased IoU scores. When considering different IoU thresholds (0.5, 0.75, 0.95), VISTAR demonstrates superior precision and recall across all levels, indicating more accurate and consistent visual grounding of the predicted objects.

\textbf{Quantitative results of textual explanations. }We then evaluate the quality of textual rationales generated by our model compared to the backbone model \textit{NVILA-8B}~\citep{Liu2024NVILAEF}. 
For \textit{NVILA-8B}~\citep{Liu2024NVILAEF}, we prompt the model with the instruction \textit{``Please explain in detail."} during inference to obtain detailed explanations. Due to the open-ended nature of these explanations, we use \textit{GPT-4-turbo}~\citep{Achiam2023GPT4TR} to assess the semantic similarity between generated explanations and ground truth answer in terms of accuracy. Additionally, to ensure a fair comparison, we evaluate the semantic similarity, operation accuracy without arguments and logical consistency of our model's predicted SoT with the ground truth annotations, also leveraging \textit{GPT-4-turbo}. \textbf{Sub-task operation accuracy} (denoted as Op. Acc) specifically measures whether each reasoning step (operation prediction) aligns with expected sub-task operations without considering detailed arguments, and \textbf{logical accuracy} (denoted as Logical Acc) evaluates if the overall reasoning process logically supports the final answer, considering the operations and intermediate results. Results in Table~\ref{tab:text_exp_result} demonstrates that VISTAR outperforms on textual interpretability across all metrics compared to \textit{NVILA-8B}~\citep{Liu2024NVILAEF}, indicating its superior capability in maintaining logical coherence throughout the reasoning process. These results confirm the effectiveness of incorporating structured SoT rationales into MLLMs for improved interpretability and logical reasoning performance.

\begin{figure}[!ht]
    \centering
    \includegraphics[width=0.5\textwidth]{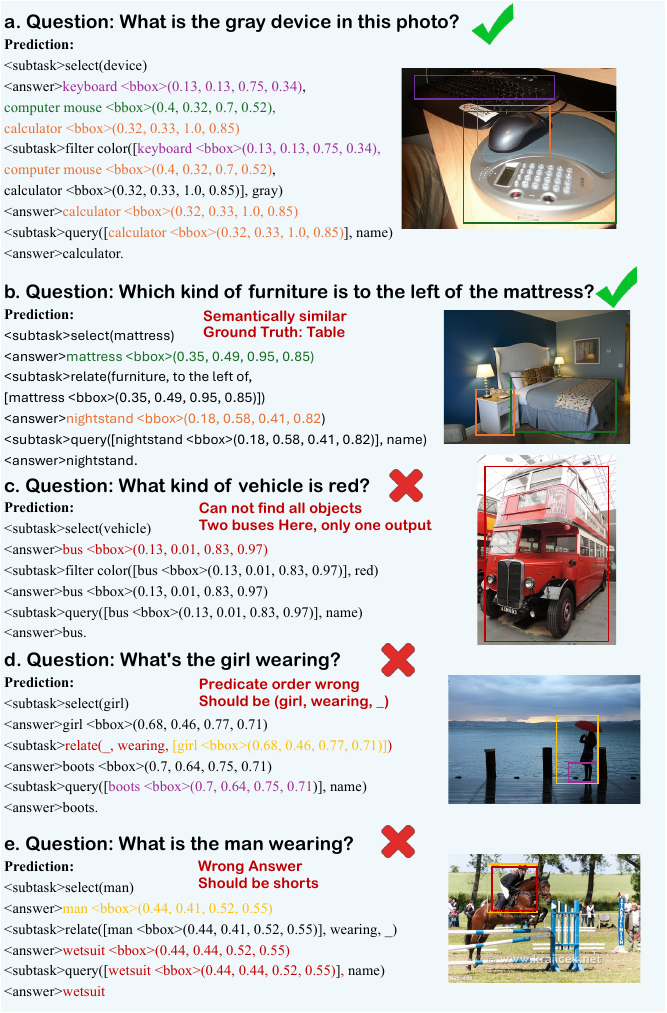} 
    \caption{\textbf{Examples of successful and failed cases on human evaluation.} We present some successful and failed cases for our \texttt{SoT} prediction based on VISTAR.}
    \label{fig:human_evaluation}
\end{figure}

\begin{figure*}[!ht]
    \centering
    \includegraphics[width=0.8\textwidth]{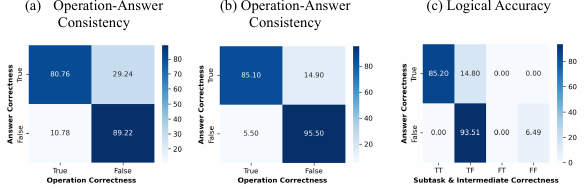} 
    \caption{\textbf{Consistency evaluation between answers and reasoning steps.} (a) Ablation results without intermediate answer supervision, evaluated by \textit{GPT-4-turbo} over operation and arguments with answers (b) Results from the full training procedure, evaluated by \textit{GPT-4-turbo} over operation and arguments with answers (c) Human evaluation results, assessing the consistency of predicted sub-tasks with their arguments and intermediate results relative to the final answers. Notably, `T' in the figure means true, 'F' means false and `TT' indicates both the sub-task and intermediate answer are correct while `TF', `FT', and `FF' represent cases where either the sub-task, intermediate answer, or both are incorrect.}
    \label{fig:consistency}
\end{figure*}

\begin{table}[h]
    \centering
    \caption{\textbf{Quantitative evaluation of textual rationales.} We compare VISTAR with \textit{NVILA-8B}~\citep{Liu2024NVILAEF} on answer accuracy (evaluated by \textit{GPT-4-turbo}~\citep{Achiam2023GPT4TR}), sub-task operation accuracy (logical correctness), and consistency of textual explanations on a sampled validation set.}
    \label{tab:text_exp_result}
    \scalebox{0.8}{
    \begin{tabular}{lccc}
        \toprule
        &Ans. Acc & Op. Acc & Logical Acc \\
        \midrule
        NVILA-8B~\citep{Liu2024NVILAEF} & 61.2  & -  & 88.9 \\
        \midrule
        \textbf{VISTAR} & \textbf{64.8} & \textbf{98.5} & \textbf{90.3} \\
        \bottomrule
    \end{tabular}}
\end{table}

\textbf{Human evaluation on SoT. }We conduct the human evaluation of predicted \texttt{SoTs} around 300 samples to present the \textbf{logical accuracy}. We assume that when the final predicted answer is correct, the correctness ratio of its sub-task operation sequences, including arguments and intermediate results, should also be high. Conversely, when the final answer is incorrect, either the sub-task operations or the intermediate answers should exhibit a high false ratio. As shown in Fig.~\ref{fig:consistency}(c), the correctness ratio of the sub-task operation sequences, including arguments and intermediate results achieves 85.2\% when the prediction is correct, which reveals the high consistency of our prediction. For the situation where the answer is true and the operation is true but the intermediate answer is wrong, it is due to the following reasons: 1) Semantically similar, the model predicts an object that is closely related to the ground truth but not identical. For instance (Fig.~\ref{fig:human_evaluation} example (b)), it identifies a nightstand instead of a table. 
2)fail to identify all relevant objects sometimes while it does not affect the final prediction, as shown in Fig.~\ref{fig:human_evaluation} example (c), the model correctly localizes one bus, but the scene contains two. 3) the wrong order of predicate, as shown in Fig.~\ref{fig:human_evaluation}, example (d), the correct reasoning should follow the structure \texttt{(girl, wearing, \_)}, but the model sometimes generates \texttt{(\_, wearing, girl)}, leading to incorrect relational understanding. The majority of incorrect predictions stem from errors in object detection. As shown in Fig.~\ref{fig:human_evaluation} example (e), the model struggles to identify fine-grained objects like shorts.


\subsection{Ablation studies}\label{subsec:ablation}
We conduct ablation studies by systematically removing different components in our SoT rationales to analyze their individual contributions to interpretability. Specifically, we evaluate the effects of (1) the inclusion of bounding boxes for visual explanations and (2) intermediate answers at each reasoning step. In terms of interpretability, we continue to perform our ablation studies on the previously mentioned sampled validation set. For the model trained without bounding boxes in SoT, we follow the instruction (see in Appendix \S\ref{sec:prompt_bbox} Fig.~\ref{fig:instruc_bbox}) to generate the bounding boxes comparing these predictions using adapted IoU. Consequently, removing the bounding boxes significantly reduces IoU scores from 48\% to 44\%. Next, we ablate the model trained without intermediate answers. For this setup, we focus on \textbf{operation-answer consistency} by examining whether operations with arguments correctly align with the final answers. Specifically, when the final answer is correct, we measure how frequently the corresponding operations with arguments are correct. Conversely, when the final answer is incorrect, we expect the operations with arguments to be predominantly incorrect. Fig.~\ref{fig:consistency}(a) shows the operation-answer consistency for the ablated model (trained without intermediate answers) and Fig.~\ref{fig:consistency}(b) shows this for our proposed model, VISTAR (trained with intermediate answers). As demonstrated in the figure, there is a decrease in consistency between the correctness of final answers and corresponding operations when intermediate answers are removed compared to our proposed method.

\section{Conclusion}
In this work, we introduce VISTAR, a framework designed to enhance the interpretability of MLLMs in VQA. By incorporating SoT rationales, VISTAR decomposes complex queries into structured reasoning steps, providing both object-level visual and textual explanations along with accurate answers. 
{However, there are certain limitations.}

VISTAR's reliance on datasets with explicit sub-task operations and scene graphs limits its applicability to unstructured datasets, which may be mitigated by generating pseudolabels for them by using predictions from a pretrained model. 
Similarly, it cannot generate unseen sub-task operations that were not included in the training. For instance, as OCR operation was not included in the GQA~\citep{Hudson2019GQAAN} dataset, it fails to perform well in OCR-based tasks on TextVQA~\citep{singh2019towards} and OCR-reliant VQAv2~\citep{balanced_vqa_v2} questions.

{
    \small
    \bibliographystyle{ieeenat_fullname}
    \bibliography{main}
}
\clearpage
\setcounter{page}{1}
\setcounter{section}{0}  
\renewcommand{\thesection}{\Alph{section}} 
\renewcommand{\thesubsection}{\Alph{section}.\arabic{subsection}}
\onecolumn





\section{Examples Outputs of Subtask-of-Thought}\label{sec:example_of_sot}
\begin{figure}[h]
    \centering
    \includegraphics[width=1.0\textwidth]{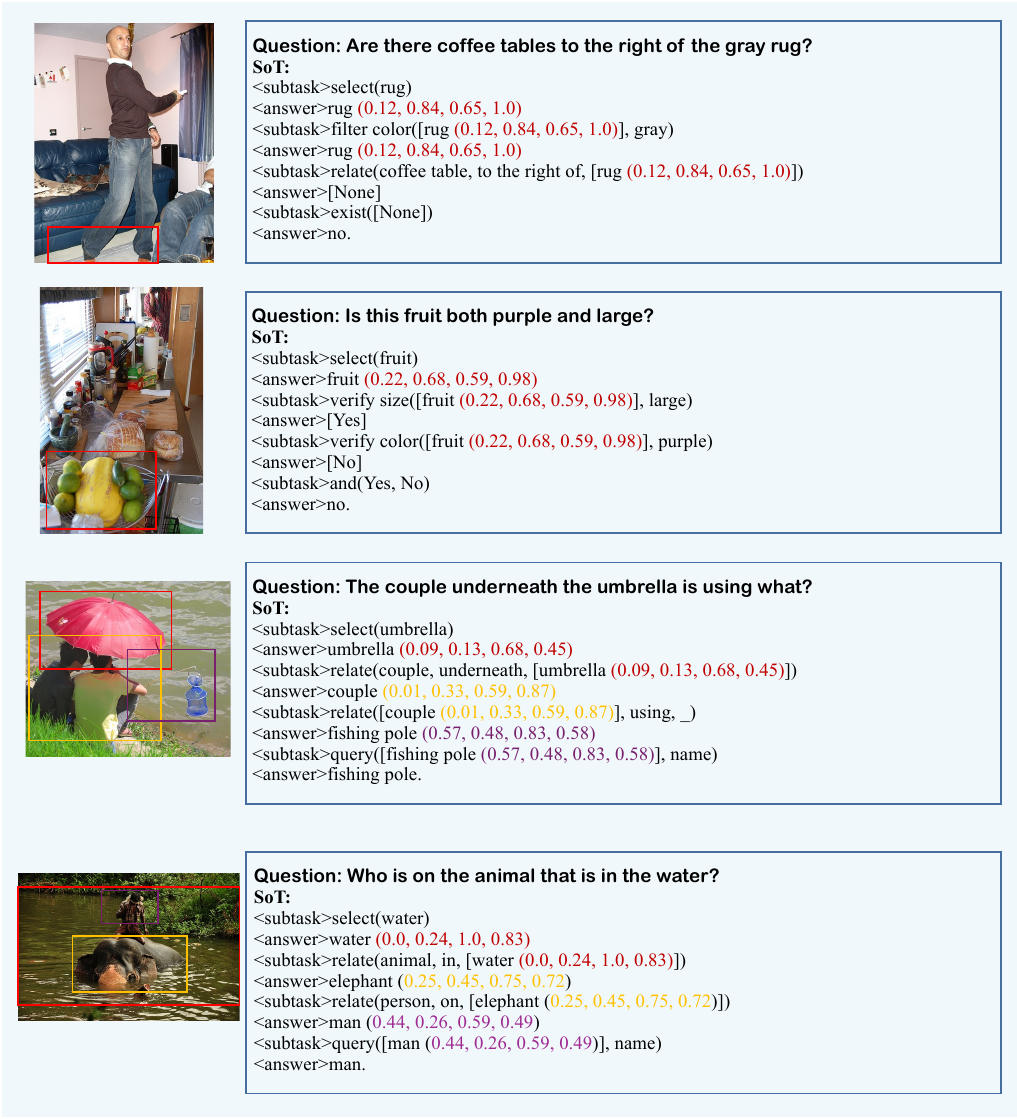} 
    \caption{\textbf{Examples of LLM-generated Subtask-of-Thought}}
    \label{fig:example_of_sot}
\end{figure}

We present some examples of LLM-generated \texttt{SoT} here, which decomposes complex tasks into sub-task and provides visual explanations, namely object-level bounding boxes, shown in Fig.~\ref{fig:example_of_sot}.


\clearpage
\section{Prompt For Subtask-of-Thought Generation}\label{sec:prompt_for_sot}
We take each query, related scene graph, sub-task operation sequences and the following prompt as input for \textit{LLaMA-3.1-70B-Instruct}~\citep{Touvron2023LLaMAOA} to generate SoT.
\begin{Verbatim}
These are all the operations possibly presented in the programs:
1. select:
{
    Input: ([Object]), 
    Output: [Object 1 (location), Object 2 (location), ...]
}
2. exist: 
{
    Input: ([Object_List]), 
    Output: [Boolean/None]
}
3. relate:
{
    Input: (Subject, Relation, Object)/(Subject, Relation)
            /(Relation, Object),
    Output: [Object 1 (location), Object 2 (location), ...]
}
4. filter: 
{
    Input: ([Object_List], Attribute), 
    Output: [Object 1 (location), Object 2 (location), ...]
}
5. verify: 
{
    Input: ([Object_List], Attribute), 
    Output: [Boolean]
},
6. verify rel: 
{   
    Input: (Subject, Relation, Object), 
    Output: [Boolean]
},
7. choose name, hposition (means horizontal position), 
   vposition (means vertical position), 
   material, location, color, place and other attribute:
{
    Input: (Object, Choice1|Choice2), 
    Output: [Right Choice]
}
8. choose rel:
{
    Input: (Subject, Choice1|Choice2)/(Choice1|Choice2, Object),
    Output: [Right Choice]
}
9. and: 
{
    Input: (Attribute, Attribute), 
    Output: [Boolean]
}
10. or: 
{
    Input: (Attribute, Attribute), 
    Output: [Boolean]
}
11. common: 
{
    Input: (Object, Object), 
    Output: [Attribute]
}
12. query: 
{
    Input: ([Object_List], Attribute), 
    Output: [Attribute]
}
13. compare: 
{
    Input: (Object, Object, Attribute), 
    Output: [Attribute]
}
14. same color, shape, material, attr: 
{   
    Input: (Object, Object), 
    Output: [Boolean]
}
15. same: 
{
    Input: ([Object_List], Attribute), 
    Output: [Boolean]
}
16. different color, shape, material, attr: 
{
    Input: (Object, Object), 
    Output: [Boolean]
}
17. different: 
{
    Input: ([Object_List], Attribute), 
    Output: [Boolean]
}
Examples:
Question: Do the bananas to the left of the plantains look large and yellow?
Operation: 
[
    {"operation": "select", "dependencies": [], "argument": "plantains (681259)"}, 
    {"operation": "relate", "dependencies": [0], 
     "argument": "bananas, to the left of s (681264)"}, 
    {"operation": "verify size", "dependencies": [1], "argument": "large"}, 
    {"operation": "verify color", "dependencies": [1], "argument": "yellow "}, 
    {"operation": "and", "dependencies": [2, 3], "argument": ""}
]
Object description: 
{
    "#1":
    {
        "id":"681253","name":"banana",
        "attributes":["small","yellow"],"location":[237,87,310,117],
        "relations":["to the left of #10"]
    },
    "#2":
    {
        "id":"681254","name":"meal",
        "attributes":[],"location":[58,121,188,232],
        "relations":[]
    },
    "#3":
    {
        "id":"681255","name":"plate",
        "attributes":["white","full"],"location":[30,111,206,249],
        "relations":["to the left of #5","of #2","with #2","near #6","to the left of #6"]
    },
    "#4":
    {
        "id":"681256","name":"spoon",
        "attributes":["large","metal","silver"],"location":[0,196,140,261],
        "relations":["on #3","to the left of #5","in #3","to the left of #6"]
    },
    "#5":
    {
        "id":"681257","name":"dish",
        "attributes":["cream colored"],"location":[187,199,295,280],
        "relations":["inside #6","to the right of #4","in #6","to the right of #3"]},
    "#6":
    {
        "id":"681258","name":"bowl",
        "attributes":["full"],"location":[178,184,293,283],
        "relations":["next to #3","of #5","near #3",
        "to the right of #4","to the right of #8","to the right of #3"]},
    "#7":
    {
        "id":"681259","name":"plantains",
        "attributes":["red"],"location":[346,0,391,70],
        "relations":["to the right of #12"]
    },
    "#8":
    {
        "id":"681260","name":"rice",
        "attributes":["piled","white"],"location":[57,162,150,219],
        "relations":["on #3","to the left of #6"]},
    "#9":
    {
        "id":"681261","name":"meat",
        "attributes":["small","brown","delicious"],"location":[68,123,92,150],
        "relations":["on #3","inside #3"]},
    "#10":
    {
        "id":"681262","name":"straw",
        "attributes":["white","plastic"],"location":[402,55,417,150],
        "relations":["to the right of #15",
        "to the right of #14","to the right of #1"]},
    "#11":
    {
        "id":"681263","name":"picnic",
        "attributes":["delicious"],"location":[0,0,499,374],
        "relations":[]},
    "#12":
    {
        "id":"681264","name":"bananas",
        "attributes":["small","yellow"],"location":[268,32,317,82],
        "relations":["to the left of #7"]},
    "#13":
    {
        "id":"681265","name":"spots",
        "attributes":[],"location":[245,92,271,108],
        "relations":[]},
    "#14":
    {
        "id":"681267","name":"banana",
        "attributes":["small","yellow"],"location":[248,55,312,89],
        "relations":["to the left of #10"]},
    "#15":
    {
        "id":"681268","name":"tablecloth",
        "attributes":["white"],"location":[0,0,396,374],
        "relations":["to the left of #10"]},
    "#16":
    {
        "id":"681269","name":"onions",
        "attributes":["green"],"location":[90,147,114,163],
        "relations":[]}
    }
Final Answer: No
Result:
[
    {
        "Operation": "select(plantains)",
        "Answer": "[#7 (346, 0, 391, 70)]"
    },
    {
        "Operation": "relate(bananas, to the left of, [#7])",
        "Answer": "[#12 (268, 32, 317, 82)]"
    },
    {
        "Operation": "verify size([#12], large)",
        "Answer": "[No]"
    },
    {
        "Operation": "verify color([#12], yellow)",
        "Answer": "[Yes]"
    },
    {
        "Operation": "and(No, Yes)",
        "Answer": "[No]"
    }
]
Question: Are there both bowls and bananas in this image?
Operation: 
[
    {"operation": "select", "dependencies": [], "argument": "banana (681253)"}, 
    {"operation": "exist", "dependencies": [0], "argument": "?"},
    {"operation": "select", "dependencies": [], 
    "argument": "bowl (681258) "}, 
    {"operation": "exist", "dependencies": [2], "argument": "?"},
    {"operation": "and", "dependencies": [1, 3], "argument": ""}
]
Final Answer: Yes

Result:
[
    {
        "Operation": "select(banana)",
        "Answer": "[#1 (237, 87, 310, 117), 
        #12 (268, 32, 317, 82), #14 (248, 55, 312, 89)]"
    },
    {
        "Operation": "exist([#1, #12, #14])",
        "Answer": "[Yes]"
    },
    {
        "Operation": "select(bowl)",
        "Answer": "[#6 (178,184,293,283)]"
    },
    {
        "Operation": "exist([#6])",
        "Answer": "[Yes]"
    },
    {
        "Operation": "and(Yes, Yes)",
        "Answer": "[Yes]"
    }
]

Note:
Operation in Response:
1. Important:
{
    "operation":"choose rel","dependencies":[0],
    "argument":"rice,to the left of|to the right of,s (681260)"
} 
represents 'rice' is the Subject (681260) of Relation 'to the left of|to the right of', 
should be 
{
    "choose rel(rice, to the left of|to the right of, ['The Object List of Relation'])".
}
{
    "operation":"verify rel","dependencies":[0],
    "argument":"bed,to the left of,o (2287058)"
} 
represents 'bed' is the Object (2287058) of Relation 'to the left of', should be 
{
    "verify rel(['The Subject List of Relation', to the left of, bed])".
}
{
    "operation": "relate", "dependencies": [0],
    "argument": "_,in,s (681257)"
} 
represents '_' is the Subject (681257) of Relation 'in', 
should be 
{
    "relate(_, in, ['The Object List of Relation'])"
}
{
    "operation": "relate", "dependencies": [0], "argument": "_,wearing,o (2029572)"
} 
represents '_' is the Object (2029572) of Relation 'wearing',
should be 
{
    "relate(['The Subject List of Relation'], wearing, _)".
}
{
    "operation":"verify rel","dependencies":[1],
    "argument":"skateboard,riding,o (-)"}
}
represents 'skateboard' is the Object of Relation 'riding', 
should be 
{
    "verify rel('The Subject List of Relation', riding, skateboard)".
}

Namely,
"s" in the argument represents current object in the argument is Subject of current Relation.
"o" in the argument represents current object in the argument is Object of current Relation.
'_' in the argument represents the placeholder.
"o" and "s" should not appear in the Operation. Ensure the argument follows the subject-relation-object order based on context.

Intermediate Answer in Response:
1. Important:
For all operations involving selection, like 'select(person)', 
Answer should include all related objects for the target, including 
singular and plural in naming or attributes, When the question or operation specifies a relationship or context, such as Operation 'relate' and 'filter', filter the selection to include only those objects that match the specified context or relation.

In cases where a verification is needed, such as checking attributes
('verify size', 'verify color'), ensure that the object being verified is relevant to the relationship or position specified in the previous steps.

Avoid including Unrelated Objects that do not fit the exact  positional or contextual requirement of the query. 
Use the attributes, locations, and relations provided in the object description to make accurate selections.

2. Important: If the Object does not exist, the answer should be [None].

3. Important: The final answer should always match the logical conclusion drawn from the operations and their reasoning steps. Noticeably, For Operation 'choose rel' in the last step, keeping identical with Final Answer. For example, choose rel(Subject, to the left of|to the right of, Object) should simply Return 'left' or 'right' based on reasoning steps.

4. Important: Answer should align with Given Relations and Locations in Object Description.

5. Important: For Operation 'select(scene)' should return 'there are' with key object IDs without bounding boxes, which are necessary to answer the question like 'there are [#2, #3, #5]' where #2, #3, #5 are key objects ID to answer the question and should be the argument for the next operation.

Based on the above example, Just Output your response in JSON format without Any Extra Explanation, keeping the same format with the given example.

\end{Verbatim}

\clearpage
\section{Sub-task Operation Definition}
\label{sec:subtask_op_def}
In Table~\ref{tab:function_1} and Table~\ref{tab:function_2}, we show all the sub-task operations used for the generation of SoT, including the name of each operation, the corresponding arguments, the function of each operation and its output.
\begin{table}[h]
    \centering
    \small
    \caption{List of sub-task operations and their descriptions including arguments, meaning and output (part one)}
    \scalebox{0.7}{
    \begin{tabular}{l|c|c|c}
        \toprule
        \textbf{Operation} & \textbf{Arguments} & \textbf{Meaning} & \textbf{Ouput} \\
        \midrule
        select & (object) & find the object & object name with bounding box\\
        \midrule
        relate & (subject, relation, object) &  find the subject or object that has the relation with object or subject & object name with its bounding box\\
        \midrule
        filter realism & \multirow{38}{*}{objects, attribute} & \multirow{38}{*}{filer the objects that have the specified ``attribute value"} & \multirow{38}{*}{object name with bounding box}\\
        filter brightness & & & \\
        filter texture & & & \\
        filter depth & & & \\
        filter weight & & & \\
        filter orientation & & & \\
        filter event & & & \\
        filter liquid & & & \\
        filter company & & & \\
        filter race & & & \\
        filter hardness & & & \\
        filter room & & & \\
        filter pattern & & & \\
        filter length & & & \\
        filter material & & & \\
        filter hposition & & & \\
        filter position & & & \\
        filter size & & & \\ 
        filter pose & & & \\
        filter activity & & & \\
        filter shape & & & \\
        filter height & & & \\
        filter age & & & \\
        filter sportActivity & & & \\
        filter face expression & & & \\
        filter length & & & \\
        filter cleanliness & & &\\
        filter sport & & & \\
        filter weather & & & \\
        filter state & & & \\
        filter thickness& & & \\
        filter opaqness& & & \\
        filter flavor& & & \\
        filter fatness& & & \\
        filter width& & & \\
        filter tone& & & \\
        filter gender& & & \\
        filter & & & \\
        \midrule
        choose weather & \multirow{27}{*}{\parbox{5cm}{\centering (subject, choice 1 $|$ choice 2) \\(choice 1 $|$ choice 2, object)}} & \multirow{27}{*}{choose which attribute value does the subject/object have}& \multirow{27}{*}{choice 1/choice 2}\\
        choose hposition & & & \\
        choose vposition & & & \\
        choose color & & & \\
        choose name & & & \\
        choose material & & & \\
        choose location & & & \\
        choose size & & & \\
        choose place & & & \\
        choose younger & & & \\
        choose older & & & \\
        choose length & & & \\
        choose pose& & & \\
        choose activity & & & \\
        choose height& & & \\
        choose less healthy& & & \\
        choose sportActivity& & & \\
        choose shape& & & \\
        choose healthier& & & \\
        choose cleanliness& & & \\
        choose state & & & \\
        choose thickness& & & \\
        choose pattern& & & \\
        choose fatness& & & \\
        choose shorter& & & \\
        choose higher& & & \\
        choose company& & & \\
        choose taller& & & \\

        \bottomrule
    \end{tabular}
    }
    
    \label{tab:function_1}
\end{table}

\begin{table}[h]
    \centering 
    \caption{List of sub-task operations and their descriptions including arguments, meaning and output (part two)}
    \scalebox{0.7}{
    \begin{tabular}{l|c|c|c}
        \toprule
        \textbf{Operation} & \textbf{Arguments} & \textbf{Meaning} & \textbf{Ouput} \\
        \midrule
        choose realism&\multirow{18}{*}{\parbox{5cm}{\centering (subject, choice 1 $|$ choice 2) \\(choice 1 $|$ choice 2, object)}} & \multirow{18}{*}{choose which attribute value does the subject/object have}& \multirow{18}{*}{choice 1/choice 2} \\
        choose larger& & & \\
        choose hardness& & & \\
        choose smaller& & & \\
        choose brightness& & & \\
        choose lower& & & \\
        choose rel & & & \\
        choose age & & & \\
        choose weight & & & \\
        choose depth& & & \\
        choose flavor& & & \\
        choose race& & & \\
        choose opaqness& & & \\
        choose gender& & & \\
        choose face expression& & & \\
        choose tone& & & \\
        choose width& & & \\
        choose & & & \\
        \midrule
        verify rel & (subject, relation, object) & verify the relation between subject and object & yes/no\\
        \midrule 
        verify state & \multirow{36}{*}{(object, attribute)}& \multirow{36}{*}{verify is the object has the attribute} & \multirow{36}{*}{yes/no}\\
        verify pose & & & \\
        verify height & & & \\
        verify location & & & \\
        verify position & & & \\ 
        verify size & & & \\
        verify material & & & \\
        verify length & & & \\
        verify weather & & & \\
        verify shape & & &\\
        verify place & & & \\
        verify pattern & & & \\
        verify cleanliness & & & \\
        verify thickness & & & \\
        verify activity & & & \\
        verify tone& & & \\
        verify hardness& & & \\
        verify face expression& & & \\
        verify age & & & \\
        verify sportActivity& & & \\
        verify width& & & \\
        verify fatness& & & \\
        verify opaqness& & & \\
        verify weight& & & \\
        verify depth& & & \\
        verify gender& & & \\
        verify company& & & \\
        verify realism& & & \\
        verify type& & & \\
        verify flavor& & & \\
        verify brightness& & & \\
        verify texture& & & \\
        verify color& & & \\
        verify race& & & \\
        verify room& & & \\
        verify & & & \\
        \midrule
        same & (objects, attribute) &if objects are of the same attributes & yes/no \\
        \midrule
        different & (objects, attribute) &if objects are of different attributes & yes/no\\
        \midrule
        same material & (object 1, object 2) &if objects 1 and object 2 are of the same material & yes/no \\
        \midrule
        same color & (object 1, object 2) &if objects 1 and object 2 are of the same color & yes/no\\
        \midrule
        different color & (object 1, object 2) &if objects 1 and object 2 are of different color &yes/no \\
        \midrule
        common & (object 1, object 2)& what do object 1 and object 2 have in common & attribute\\
        \midrule
        and  & (attribute 1, attribute 2) & if both attribute 1 and attribute 2 are true & yes/no\\
        \midrule
        or & (attribute 1, attribute 2) & if either attribute 1 and attribute 2 are true & yes/no \\


        \bottomrule
    \end{tabular}
    }
    
    \label{tab:function_2}
\end{table}

\clearpage
\section{Prompt for the output of bounding boxes}\label{sec:prompt_bbox}
For the backbone model \textit{NVILA-8B}~\citep{Liu2024NVILAEF} and the ablation model that removes bounding box, we need to compare with our visual explanation. Hence, we prompt these models during training using the following prompt, as shown in Fig.~\ref{fig:instruc_bbox}
\begin{figure}[h]
    \centering
    \begin{tcolorbox}[colback=gray!30, arc=3mm, boxrule=1pt]
    \small
    \#\#\# For the prediction of the final answer
    
    Question: [Q]
    
    Answer the question using a single word or phrase.
    
    Response: [X]\\
    
    \#\#\# For the prediction of bounding box
    
    Question: [Q]
    
    Answer: [X]
    
    Provide the bounding box coordinate of the region [X].
    \end{tcolorbox}
    \caption{\textbf{Instruction for bounding box prediction during inference}}
    \label{fig:instruc_bbox}
    
\end{figure}

\section{Example of converting sub-task operations into subtask-of-thought}\label{sec:example_of_sot_op}
In the GQA~\citep{Hudson2019GQAAN} dataset, they provide sub-task operation sequences, consisting of the sub-task operation and the corresponding arguments. We reformat the sub-task operation sequences by restructuring the format of each operation and its arguments and providing intermediate results after each operation.
\begin{figure*}[htbp]
    \centering
\begin{tcolorbox}[colback=myblue_fill!30, colframe=myblue_shape, arc=3mm, boxrule=1pt]

    \#\#\# \textbf{Question}\\
    \textit{What is the} \textbf{\textcolor{mygold}{name}} \textit{of the item of} \textbf{\textcolor{mygold}{furniture}} \textit{that has the} \textcolor{myblue}{same color} \textit{as the} \textcolor{mypink}{curtain} \textit{that is} \textcolor{mybrown}{to the right} \textit{of the} \textcolor{mygreen}{garland}?

    \vspace{0.2cm}
    \#\#\# \textbf{Sub-task Operation Sequence}\\
    \textbf{Select:} \textbf{\textcolor{mygreen}{garland}}$\rightarrow$
    \textbf{Relate:} \textcolor{mypink}{curtain}, \textcolor{mybrown}{to the right}, \textcolor{mygreen}{garland}$\rightarrow$ 
    \textbf{Relate:} \textcolor{mygold}{furniture}, \textcolor{myblue}{same color}, \textcolor{mypink}{curtain} $\rightarrow$ 
    \textbf{Query:} \textcolor{mygold}{name}
    
    \vspace{0.2cm}

    \#\#\# \textbf{Subtask-of-Thought}\\
    \textless \textit{subtask}\textgreater \textit{select(\textbf{\textcolor{mygreen}{garland}})\textless answer\textgreater \textbf{\textcolor{mygreen}{garland}} \textless bbox\textgreater (0.51, 0.0, 0.54, 0.09)}\\
    \textless \textit{subtask}\textgreater
    \textit{relate(\textbf{\textcolor{mypink}{curtain}, \textcolor{mybrown}{to the right of}, \texttt{[}\textcolor{mygreen}{garland}} \textless bbox\textgreater (0.51, 0.0, 0.54, 0.09)\textbf{\texttt{]}})\textless answer\textgreater \textbf{\textcolor{mypink}{curtain}}} \textless bbox\textgreater (0.73, 0.0, 0.87, 0.58)\\
    \textless \textit{subtask}\textgreater
    \textit{relate(\textbf{\textcolor{mygold}{furniture}, \textcolor{myblue}{same color}, \texttt{[}\textcolor{mypink}{curtain}} \textless bbox\textgreater (0.73, 0.0, 0.87, 0.58)\textbf{\texttt{]}})\textless answer\textgreater \textbf{\textcolor{mygold}{coach}}} \textless bbox\textgreater (0.12, 0.48, 0.71, 0.97)\\
    \textless \textit{subtask} \textgreater
\textit{query(\textbf{\texttt{[}\textcolor{mygold}{couch}} \textless bbox\textgreater(0.12, 0.48, 0.71, 0.97)\textbf{\texttt{]}}, name)}\textless answer\textgreater \textbf{\textit{\textcolor{mygold}{coach}}}

\end{tcolorbox}
\caption{\textbf{Example of sub-task steps in GQA~\citep{Hudson2019GQAAN} dataset and our subtask-of-thought formate. }For the operations of the given query, it means first we need to identify the garland in the image, and then find the curtain positioned to the right of the garland. Next, we can locate furniture that has the same color as the curtain. Finally, retrieving the name of the identified furniture is the end. Converting from sub-task operation sequences to our SoT requires the question, scene graph, and ground truth answer, prompt(see in Appendix \S~\ref{sec:prompt_for_sot}.}
    \label{fig:subtask_example}
\end{figure*}

\clearpage

\twocolumn

\end{document}